\documentclass[letterpaper]{article} 
\usepackage{aaai2026}  
\usepackage{times}  
\usepackage{helvet}  
\usepackage{courier}  
\usepackage[hyphens]{url}  
\usepackage{graphicx} 
\usepackage{natbib}  
\usepackage{caption} 
\usepackage{algorithm}
\usepackage{algorithmic}

\usepackage{newfloat}
\usepackage{listings}
\DeclareCaptionStyle{ruled}{labelfont=normalfont,labelsep=colon,strut=off} 
\floatstyle{ruled}
\newfloat{listing}{tb}{lst}{}
\floatname{listing}{Listing}
\title{PressTrack-HMR: Pressure-Based Top-Down Multi-Person Global\\ Human Mesh Recovery}
\author{
    Jiayue Yuan, Fangting Xie, Guangwen Ouyang, Changhai Ma, Ziyu Wu, Heyu Ding, Quan Wan, Yi Ke, Yuchen Wu, Xiaohui Cai\thanks{Corresponding author.}
}
\affiliations{

    University of Science and Technology of China\\
    yuanjiayue@mail.ustc.edu.cn, caixiaohui@ustc.edu.cn
}

\usepackage{bibentry}

\usepackage{amsmath} 
\usepackage{amsfonts} 
\usepackage{amssymb}  
\usepackage{multirow}

\begin{document}

\maketitle

\begin{abstract}
Multi-person global human mesh recovery (HMR) is crucial for understanding crowd dynamics and interactions. Traditional vision-based HMR methods sometimes face limitations in real-world scenarios due to mutual occlusions, insufficient lighting, and privacy concerns. Human-floor tactile interactions offer an occlusion-free and privacy-friendly alternative for capturing human motion.
Existing research indicates that pressure signals acquired from tactile mats can effectively estimate human pose in single-person scenarios. However, when multiple individuals walk randomly on the mat simultaneously, how to distinguish intermingled pressure signals generated by different persons and subsequently acquire individual temporal pressure data remains a pending challenge for extending pressure-based HMR to the multi-person situation. In this paper, we present \textbf{PressTrack-HMR}, a top-down pipeline that recovers multi-person global human meshes solely from pressure signals. This pipeline leverages a tracking-by-detection strategy to first identify and segment each individual's pressure signal from the raw pressure data, and subsequently performs HMR for each extracted individual signal. Furthermore, we build a multi-person interaction pressure dataset \textbf{MIP}, which facilitates further research into pressure-based human motion analysis in multi-person scenarios. 
Experimental results demonstrate that our method excels in multi-person HMR using pressure data, with 89.2~$mm$ MPJPE and 112.6~$mm$ WA-MPJPE$_{100}$, and these showcase the potential of tactile mats for ubiquitous, privacy-preserving multi-person action recognition.
Our dataset \& code are available at \url{https://github.com/Jiayue-Yuan/PressTrack-HMR}.
\end{abstract}


\section{Introduction}
Multi-person global human mesh recovery (HMR) is a pivotal technology that provides comprehensive crowd dynamics information, including individual poses and trajectories, which offers significant value across various applications. For instance, it enables accurate multi-person motion capture for virtual reality, facilitates health monitoring and fall prediction in assisted living, supports detailed human movement analysis in sports and rehabilitation, and provides precise crowd behavior analysis for public safety and event management.

\begin{figure}[t]
\centering
\includegraphics[width=1\columnwidth]{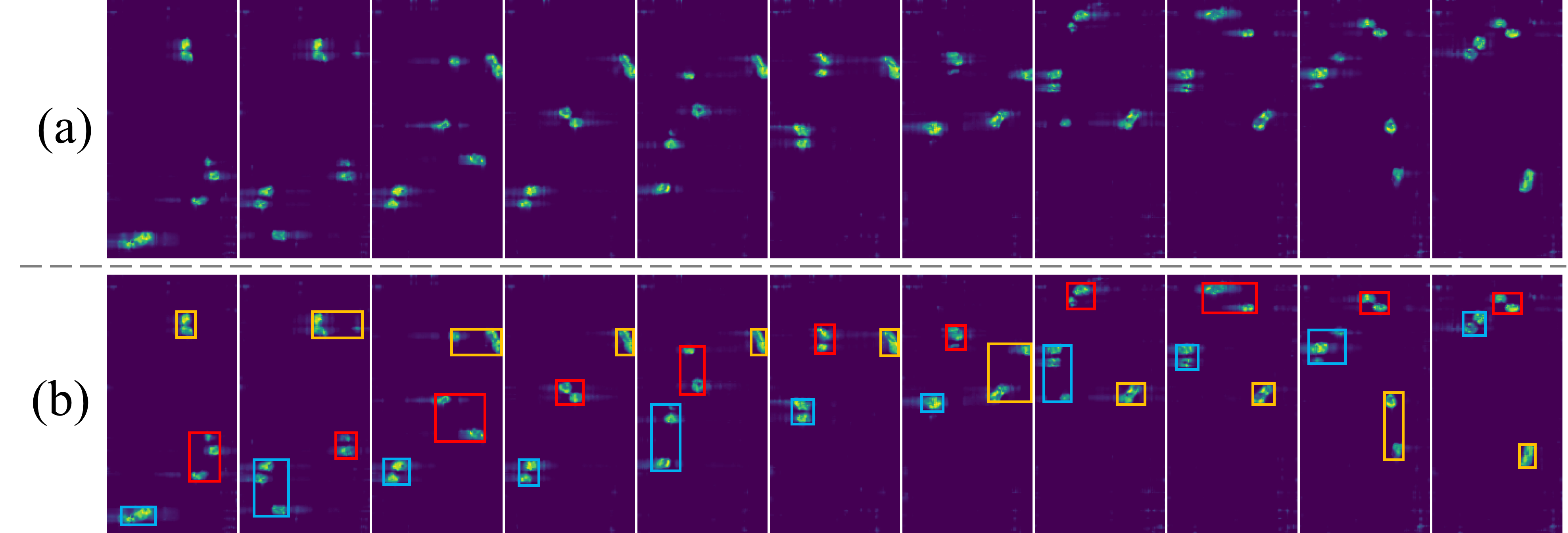} 
\caption{(a) A sequence of consecutive raw pressure maps, displayed from left to right at 0.27-second intervals. (b) The same pressure maps with different colored detection boxes representing the footprints of different individuals.}
\label{pressure_map}
\end{figure}

However, state-of-the-art RGB-based human mesh recovery methods face additional challenges in real-world crowd monitoring scenarios: (1) In crowded environments, inevitable mutual occlusion among pedestrians limits the completeness of single-view information, while deploying multiple cameras typically involves higher costs and complexity; (2) Insufficient lighting degrades visual data quality and availability in certain contexts; (3) Growing public demand for privacy protection renders camera-based solutions less desirable or even unacceptable in sensitive environments like homes, hospitals, or public training rooms. These limitations constrain the widespread application of existing RGB-based HMR methods.

We observe that tactile interactions between humans and the ground provide rich pressure information, which inherently avoids the aforementioned problems. Specifically, pressure data is robust to mutual occlusions and insufficient lighting conditions, and it intrinsically offers privacy-preserving advantages. While prior works~\cite{Luo_IntelligentCarpet, cavatar, Ren_MotionPRO} utilized mat pressure data for human pose estimation, they primarily focused on single-person scenarios.

In multi-person mobility scenarios, high pedestrian density is common, and individual movement patterns exhibit diversity and complexity.
The primary challenges for multi-person global HMR using pressure mat can be summarized as follows: 
(1) \textbf{Intra-frame individual pressure signal separation}.
When the pressure-sensing mat is occupied by multiple users simultaneously, their random stances result in intermingled pressure signals, as shown in Fig.~\ref{pressure_map}(a), which necessitates distinguishing between the signals of different individuals.
(2) \textbf{Inter-frame individual pressure signal association}.
Recent pressure-based HMR studies have demonstrated that leveraging temporal consistency is more helpful in achieving long-term stability~\cite{pimesh}. Therefore, to obtain more accurate and smoother results, it is essential to identify the continuous pressure signals belonging to the same person across successive time points, as Fig.~\ref{pressure_map}(b). 

In this work, we analyze the dynamic characteristics of pressure footprints and present \textbf{PressTrack-HMR}, a multi-person pressure-based global human mesh recovery pipeline that uses a top-down approach and consists of a PressTrack module and an HMR module.
In the PressTrack module, which employs a tracking-by-detection strategy, we first fine-tuned a YOLOv11 model~\cite{2024yolov11} as a detector to directly obtain footprint bounding boxes for different individuals from the raw pressure data. For the subsequent tracking step, we proposed a UoE-based (Union over Enclosure) similarity metric to associate pressure signals belonging to the same person across successive frames, thereby tracking and extracting continuous pressure signals for each individual.
Finally, in the HMR module, individual pressure signals were processed by a pose estimation model to generate meshes.

Although there have been some pioneering datasets with pressure, such as MoYo~\cite{moyo} and MotionPRO~\cite{Ren_MotionPRO}, they only cover single-person scenarios. While the study by~\cite{Luo_IntelligentCarpet} includes a limited number of dual-person scenarios, the relative positions of individuals in these scenarios remain almost unchanged, lacking the dynamic person-to-person interaction that is common in real-world settings. Addressing these limitations, we collected a \textbf{M}ulti-person \textbf{I}nteraction \textbf{P}ressure dataset named \textbf{MIP}. This dataset comprises data from 20 volunteers, including 20 sets of single-person data, 30 sets of multi-person data with randomly combined 2 or 3 individuals, and 2 sets of couple dance sequence, totaling over 138K synchronized tactile and visual frames.

Our experimental results show the approach's capability to generate plausible multi-person human mesh and global trajectory. Specifically, for evaluating the extraction of temporal pressure signals for each individual, we employed CLEARMOT metrics~\cite{CLEARMetrics}, achieving 93.6\% MOTA and 94.8\% MOTP. In terms of overall human mesh recovery performance, compared to ground truth derived from multi-view visual information, the pipeline exhibits a mean per joint position error (MPJPE) of 89.2~$mm$, and a whole-segment aligned MPJPE for 100-frame sequences (WA-MPJPE$_{100}$) of 112.6~$mm$.
Additionally, we perform ablation studies to assess the significance of individual components in our model and evaluate its generalization performance on unseen individuals. 
Our contributions are summarized as follows:
\begin{itemize}
    \item We propose PressTrack, a robust tracking-by-detection method specifically designed for pressure footprint. 
    \item We present PressTrack-HMR, the first pipeline to the best of our knowledge, that enables multi-person global human mesh recovery exclusively from pressure data. 
    \item We collected a multi-person interaction pressure dataset MIP, facilitating further research into pressure-based human motion analysis in multi-person scenarios.
\end{itemize}

\begin{table*}[htbp]
    \centering
        \begin{tabular}{lccccccc}
            \hline
            Datasets & Vision & Sensor & Human Body & Subject & Frames & Scenarios & interaction \\
            \hline
            I.C. \cite{Luo_IntelligentCarpet} & D.V. RGB & 96*96 & Skeleton & 10 & 180K & 1-, 2-person & -\\
            MoYo \cite{moyo} & M.V. RGB & 37*110 & SMPL & 1 & 560K & 1-person & -\\
            MotionPRO \cite{Ren_MotionPRO} & M.V. RGB & 120*160 & SMPL & 70 & 12.4M & 1-person & -\\
            \hline
            \textbf{Ours} & M.V. RGB & 120*240 & SMPL & 20 & 138K & Multi-person & $\checkmark$\\
            \hline
        \end{tabular}
    \caption{Comparison of existing pressure mat datasets. I.C.: Intelligent Carpet, D.V.: Dual-View, M.V.: Multi-View.}
    \label{dataset}
\end{table*}

\section{Related Works}
\subsection{Multi-Object Tracking (MOT)}
Object detection is the basis of MOT with numerous popular detectors available. Among these, the YOLO series detectors~\cite{yolov3, yolov7, 2024yolov11} are adopted by a large number of methods~\cite{transmot, Cheng2024YOLOWorld} for their excellent balance of accuracy and speed.
Tracking-by-detection methods directly utilize detection boxes from single images for tracking, with data association as their core. This involves computing similarity between tracklets and detections and then matching them. 
SORT~\cite{SORT} combines location and motion cues by adopting a Kalman Filter~\cite{Kalman} to predict tracklets, followed by IoU-based similarity computation. 
ByteTrack~\cite{ByteTrack} refines the association by considering both high-score and low-score detection boxes through a two-stage matching strategy, effectively retaining occluded objects often discarded by conventional methods. 
BoT-SORT~\cite{aharon2022bot} further elevates tracking performance by integrating motion model and  Re-ID feature with highly efficient association strategy.
The primary challenge in visual multi-object tracking lies in visual occlusion, and the sizes and positions of their detection boxes typically change continuously over time. Consequently, these vision-based tracking methods are not directly applicable to pressure footprint tracking.

\subsection{Vision-Based Human Mesh Recovery} 
Pioneering vision-based human mesh recovery studies mainly used pre-trained parametric models like SMPL \cite{smpl} to represent human body mesh, with inference done via optimization or regression. Optimization-based methods mainly focus on fitting the SMPL parameters to image cues such as 2D joints~\cite{smplify, exploiting, pavlakos2019expressive}.
Regression-based methods directly regress SMPL parameters under visual supervision~\cite{hmr}. 
Additionally, researchers have increasingly focused on global trajectory estimation. 
GLAMR~\cite{yuan2022glamr} uses local poses and human relationships without considering moving camera positions; 
WHAM~\cite{shin2024wham} infers motion with contact labels from foot velocity;
PromptHMR~\cite{PromptHMR} leverages multimodal prompts and full-image processing for enhanced accuracy and robustness in challenging scenarios.
Despite these advancements, occlusion and variable lighting conditions are inevitable during daily activities, which may hinder vision-based methods. Moreover, the use of cameras also raises privacy concerns, particularly in health monitoring and sports training.

\subsection{Pressure-Based Human Mesh Recovery} 
Pressure-based methods are immune to occlusion and do not require direct image or video capture, thus offering privacy protection.
Existing studies have explored applications in human mesh recovery using typical pressure sensing devices(e.g., mats~\cite{Luo_IntelligentCarpet, cavatar}, clothes~\cite{Zhang2024Garment, Zhou2023Mocapose}, bedsheets~\cite{Clever2020Bodies, pimesh}, and shoes~\cite{Zhang2024MMVP, VanWouwe2024Diffusionposer}). 
For instance, PID-HMR~\cite{wan2025bed} and PI-HMR~\cite{wu2025pihmr} utilize multi-frame pressure information as input for in-bed HMR; VP-MoCap~\cite{Zhang2024MMVP} leverages foot contact constraints to assist in defining the relative position between humans and the environment; MotionPRO~\cite{Ren_MotionPRO} proofs that pressure could provide accurate global trajectory and plausible lower body pose.
However, existing research is mainly limited to single-person scenarios, and the application of pressure sensing in multi-person scenarios remains to be explored.

\section{Dataset}
Existing datasets (as shown in Table~\ref{dataset}) predominantly feature single-person scenarios, only Intelligent Carpet~\cite{Luo_IntelligentCarpet} offers a limited number of dual-person scenarios without  interaction. To investigate pressure-based multi-person HMR, we have collected the multi-person interaction pressure (MIP) dataset. This dataset is designed to simulate realistic multi-person mobility scenarios by including the following key characteristics: (1) Concurrent presence of multiple participants; (2) Dynamic changes in relative positions; (3) Authentic interactive behaviors, including common human interaction patterns such such as mutual avoidance and following.

\textbf{Setup}
We set up a data acquisition site equipped with seven synchronized RGB cameras and a self-made pressure mat. This pressure mat (1.20$\times$2.40$m^2$, with 120$\times$240 sensing units and a pitch of 1$cm\times$1$cm$) provides a resolution superior to all other publicly available pressure datasets. The RGB cameras are used to capture body motion videos from multiple perspectives. 

\textbf{Procedures}
Twenty volunteers were recruited to take part in the experiment (12 males and 8 females, height: 1.52-1.83$m$, weight: 39-87$kg$, age: 20-27), ensuring variation and generalizability. All participants consented to the use of their data for academic purposes. We collected 20 sets of single-person data from all participants, 30 sets of multi-person data from groups of 2- or 3-person formed randomly, alongside 2 sets of couple dance sequences. Each set lasted approximately 3 minutes, totaling over 138K frames. 
During the collection procedure, volunteers performed a diverse set of individual movements, including dynamic forward walking, backward walking, standing still, and various relaxed postures. 
For multi-person groups, tasks were specifically designed to facilitate realistic interaction behaviors such as walking side-by-side, following, and walking towards each other. And couple dance groups further augment the dataset's interactivity. 
Additionally, volunteers were also allowed to  briefly step outside the pressure mat's range and then step back in, as we expect our algorithm to accommodate such changes in the number of people on the mat.

\textbf{Annotation Acquisition}
We employ the SMPL model~\cite{smpl} as the human body representation, with its ground truth parameters derived from multi-view RGB videos using Easymocap~\cite{easymocap}.
Then we re-checked the data labels and manually corrected generation errors caused by occlusions and other factors. 

\section{Pressure Pattern Analysis}

\begin{figure}[ht]
\centering
\includegraphics[width=0.95\columnwidth]{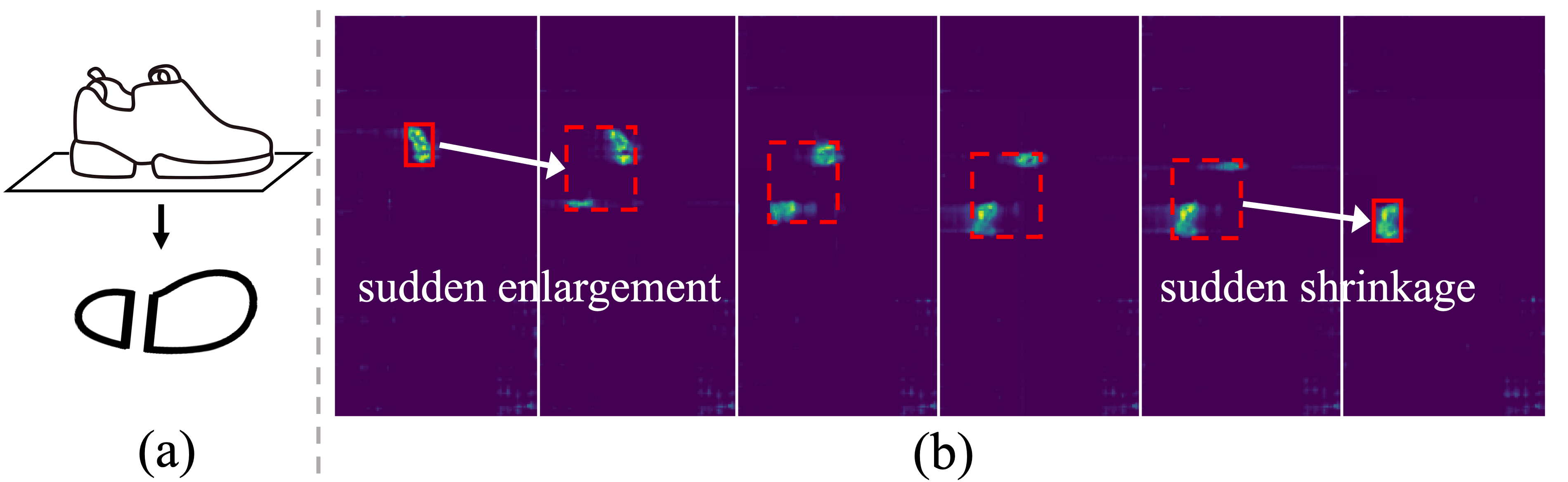} 
\caption{(a) Two separated footprint regions caused by shoe sole. (b) Dynamics of pressure maps during human locomotion. Dashed large boxes indicate a two-footed contact scenario, solid small boxes indicate single-footed contact.}
\label{characteristic}
\end{figure}

\begin{figure*}[t]
\centering
\includegraphics[width=0.95\textwidth]{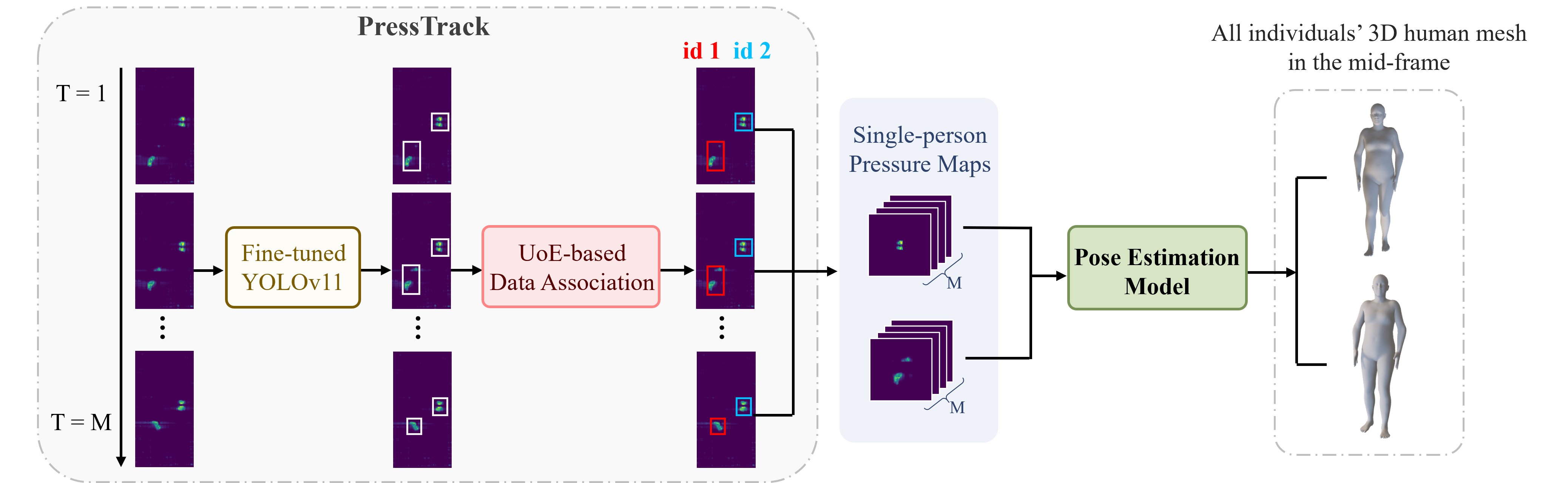} 
\caption{The framework of our proposed pipeline PressTrack-HMR.}
\label{pipeline}
\end{figure*}

As shown in Fig.~\ref{characteristic}(a), for a single foot, even when both the forefoot and hindfoot are in contact with the ground, the presence of the foot arch and the design of shoe soles can sometimes result in two smaller, disconnected pressure regions. Furthermore, during human locomotion, as the left and right feet alternately lift off the ground, the pressure signal from a single foot is inherently intermittent. 
To address these issues and ensure a continuous trajectory for an individual, it is essential to integrate all pressure signals belonging to that person, including both feet and any disconnected regions within a single foot,
Therefore, one solution is using detection boxes to contain all pressure signals belonging to the same person. 

Dynamics of pressure maps during human locomotion are depicted in Fig.~\ref{characteristic}(b). when both feet are simultaneously on the ground, the footprint detection box on the mat pressure map is a large bounding box covering both feet. As a person initiates walking, one foot (e.g., the left) first lifts off the ground, with pressure focusing onto the other foot.  At this point, the bounding box shrinks to a small size, covering only one foot. The left foot then advances a certain distance before making contact and landing again. Both feet simultaneously touch the ground once more, causing the bounding box to expand back to a large size covering both feet. Subsequently, the right foot will replicate this action of lifting off and moving forward, thereby completing the continuous locomotion. 
Based on the observations above, we can deduce two characteristics of pressure footprints regarding tracking: 
\begin{itemize}
\item Abrupt Size Changes: The alternation between single and double foot contact causes sudden changes in detection boxes' size. This reduces the reliability of the Intersection over Union (IoU) values between adjacent frames; 
\item Jump-like discontinuous Motion: The movement of the detection boxes appears discontinuous and abrupt. This renders smooth-motion-prediction-methods like the Kalman Filter ineffective for this type of data.
\end{itemize}

\section{Pipeline}
Based on the MIP dataset, we develop a pipeline for multi-person global human mesh recovery, which we call PressTrack-HMR. 
As shown in Fig.~\ref{pipeline}, our pipeline comprises two main components:
1) PressTrack: A tracking-by-detection module that extracts temporal single-person pressure maps from raw signals; 
2) Human Mesh Recovery: A deep learning module that reconstructs global human meshes from the temporal single-person pressure maps, ultimately yielding the trajectories of members in the group.

\subsection{PressTrack Module}
In this section, we implement the identification and extraction of each individual's temporal pressure signal from the raw pressure data. The core challenges are twofold: 1) Intra-frame Footprint Detection: detecting and identifying all pressure signals belonging to a single individual within one frame; 2) Inter-frame Footprint Association: accurately associating the pressure signals of the same individual across consecutive frames. Inspired by common practices in multi-object tracking within computer vision~\cite{SORT, DeepSORT, ByteTrack}, we utilize a tracking-by-detection paradigm. 

\begin{figure}[t]
\centering
\includegraphics[width=0.9\columnwidth]{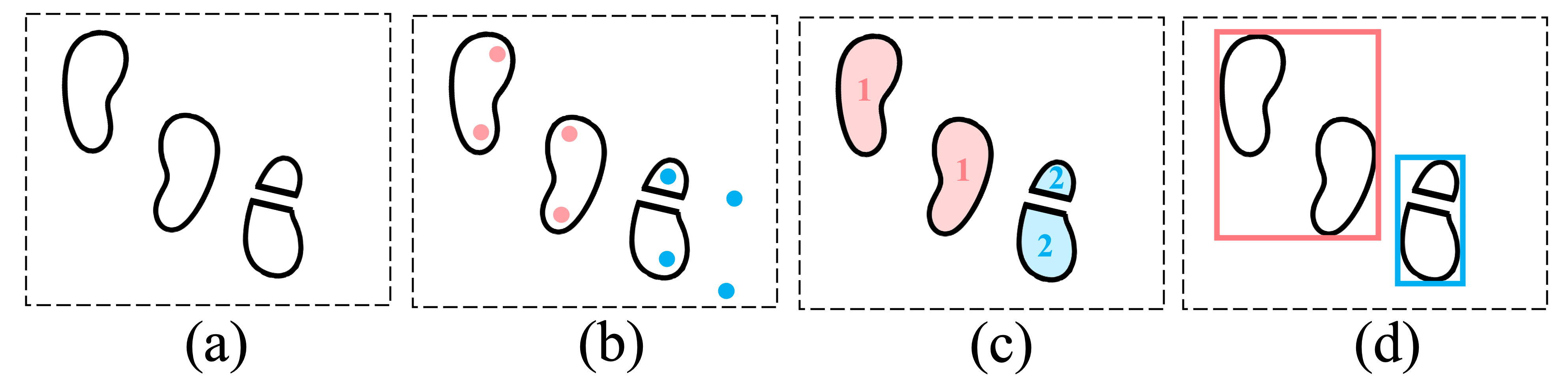} 
\caption{Object detection label generation process. (a) Initial discrete pressure regions; (b) Project 2D toe-base and ankle joints for different individuals; (c) Assign discrete regions to individuals based on geometric proximity; (d) Merge regions assigned to the same individual into one box.}
\label{label}
\end{figure}

\subsubsection{Intra-frame Footprint Detection}
We fine-tune a pre-trained YOLOv11~\cite{2024yolov11}, a cutting-edge, lightweight object detection model, on our dataset for the specific task of pressure footprint detection. 
The bounding box labels are automatically generated via a series of processes, as shown in Fig.~\ref{label}. Firstly, we employ OpenCV's thresholding method to extract initial discrete pressure regions from raw data. 
Next, for each individual, we extract the foot joint points (specifically the left and right toe-base and ankle) from the ground-truth 3D joint coordinates obtained from RGB data. The foot joint points for individual~$i$ are then projected onto the 2D coordinate system of the pressure mat to form a set of projected foot joint references~$F_i$.
For each discrete region~$r_j$ detected by OpenCV in frame $k$, with center coordinates~$c_j$, we assign an individual ID to it. This assignment is based on the geometric closest distance between the region's center and any of the projected foot joint references of an individual. Specifically, we assign the ID of individual~$i$ to region~$r_j$ as follows: 
\begin{equation}
    \text{ID}(r_j) = \arg\min_{i \in \{1, \ldots, N\}} \min_{f \in F_i} \|c_j - f\|_2
\end{equation}
where $N$ is the total number of individuals in the frame, and $\|\cdot\|_2$ denotes the Euclidean distance. Then, all regions assigned to the same ID are merged into a single box, which is the minimum bounding rectangle, to serve as the final training label for the YOLOv11 model. Finally, we manually reviewed all generated labels to ensure their quality.

\subsubsection{Inter-frame Footprint Association}
The accurate association of detected pressure footprints across consecutive frames is fundamental to extracting stable individual temporal pressure signals. However, pressure footprints exhibit dynamic characteristics different from 2D pedestrian tracking, which cause off-the-shelf visual trackers to perform poorly. 

For the task of pressure footprint tracking, we propose a simplified approach for inter-frame data association.
Rather than employing motion prediction—which in conventional tracking methods uses the target’s previous position and velocity to predict its location in the current frame, thereby facilitating more accurate association between tracks and detections—we directly compute the assignment cost matrix between detection boxes in adjacent frames. 
This cost matrix is based on the Union over Enclosure (UoE) distance, which serves as our similarity metric. UoE is proposed based on the principle that pressure footprints from different individuals do not overlap simultaneously, defined as: 
\begin{equation}
    UoE = \frac{\left | A\cup B \right |}{\left | C \right |}
\end{equation}
where A and B represent the detection boxes in the current and previous frames, C is the minimum bounding rectangle enclosing both A and B, and $|\cdot|$ denotes the area of the enclosed region. 
Additionally, detection confidence scores are integrated to refine the cost matrix, ensuring that high-confidence detections are more likely to be matched. 
The optimal assignment between detections and existing trajectories is then found using the Hungarian algorithm~\cite{Hungarian}, resulting in matched pairs for tracking. 

For robust track updates and management, we employ specific strategies to handle unmatched detections and existing tracks. Unmatched detections are processed according to their confidence scores: low-confidence detections are discarded, while those with sufficient confidence are initialized as new tracks—for example, when new individuals enter the mat’s sensing area. Unmatched tracks are temporarily marked as 'lost' and retained for a certain number of frames in anticipation of the object's reappearance, such as during a brief two-footed jump, after which the person's track can be re-associated. If a track remains unmatched for an extended period, it is considered to have exited the scene and is consequently removed. 

Finally, for each sequence of pressure footprints with the same identity, we extract and standardize the pressure data. Specifically, the data within each detection box is extracted and padded to $128 \times 128$. The center coordinates of the detection boxes serve as the localization information for the footprints on the entire pressure mat. The resulting $128 \times 128$ pressure image sequence then constitutes our final temporal single-person pressure maps. 

\subsection{Human Mesh Recovery Module}
\begin{figure}[t]
\centering
\includegraphics[width=0.9\columnwidth]{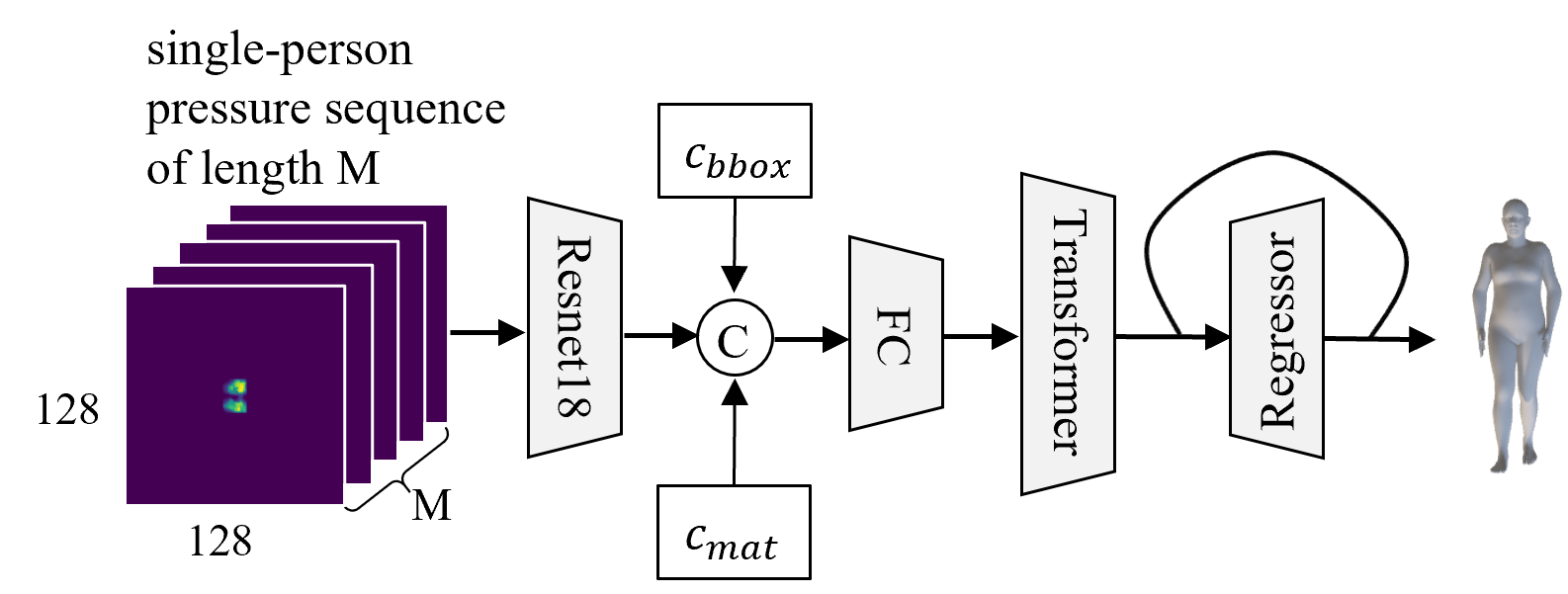} 
\caption{Architecture of our human pose estimation model.}
\label{HMRmodel}
\end{figure}
As shown in Fig.~\ref{HMRmodel}, our human pose estimation model mainly consists of three parts: image encoder, temporal encoder, and SMPL regressor. 

\textbf{Image Encoder}
We observe that the valuable information within pressure maps is encompassed by all the detection boxes (i.e., the pressure distribution within the boxes, their center coordinates, height and width). Consequently, instead of feeding the complete pressure maps directly into the image encoder, we input single-person pressure maps extracted from the detection boxes. By encoding these maps using a ResNet network\cite{resnet}, we can effectively extract image features. These image features are then concatenated with their corresponding detection boxes center coordinates~$c_{bbox}$, which denote the person's localization on the mat, and the spatial coordinates of the four mat corners~$c_{mat}$, serving as prior spatial information. These combined tensors are denoted as static features~$X = \{x_t\}_{t=1}^{T}$. A fully connected layer then adjusts its dimensionality to match the input dimension of the Transformer encoder.

\textbf{Temporal Encoder}
To harness the temporal dependencies within our static features~$X$, we utilize a Transformer model\cite{vaswani2017attention}, renowned for its proficiency in processing such data. Specifically, we compute their corresponding temporal features~$Z = \{z_t\}_{t=1}^{T}$ by employing a two-layer Transformer encoder block with position embedding, and each Transformer encoder block comprises a multi-head attention and a feed-forward layer. 

\textbf{SMPL Regressor}
Our model follows an N-to-1 mapping paradigm to estimate a single-frame human mesh from an input frame sequence. Specifically, we compute the mean feature of  Z as the integrated feature for the mid-frame. This feature is then processed by an iterative SMPL regressor~\cite{hmr}, which employed an iterative error feedback (IEF) loop strategy to iteratively regress the SMPL parameters, including pose~$\boldsymbol{\theta}$, shape~$\boldsymbol{\beta}$, and global translation~$T$. Ultimately, we obtain the human mesh information for the target mid-frame.

\textbf{Supervision}
The overall loss function of our model can be expressed as follows:
\begin{equation}
    \mathcal{L} = \mathcal{L}_{\text{SMPL}} + \mathcal{L}_{3D}
\end{equation}
where~$\mathcal{L}_{\text{SMPL}}$ presents the mean absolute error (MAE) between the estimated SMPL parameters and ground truths,~$\mathcal{L}_{3D}$ minimizes L2 loss between estimated and ground truth 3D joints regressed from SMPL vertices. Each term is calculated as:
\begin{equation}
    \mathcal{L}_{\text{SMPL}} = \lambda_{\beta}  |\hat{\beta} - \beta| + \lambda_{\theta} |\hat{\theta} - \theta| + \lambda_{T} |\hat{T} - T|
\end{equation}

\begin{equation}
    \mathcal{L}_{3D} = \lambda_{3D} \|J_{3D} - \hat{J}_{3D}\|_2
\end{equation}
where~$\hat{x}$ represents the ground truth for the corresponding estimated variable~$x$, and~$\lambda_{\beta}$,~$\lambda_{\theta}$,~$\lambda_{T}$,~$\lambda_{3D}$ are the weighting coefficients for each loss component.

\begin{figure*}[t]
\centering
\includegraphics[width=0.9\textwidth]{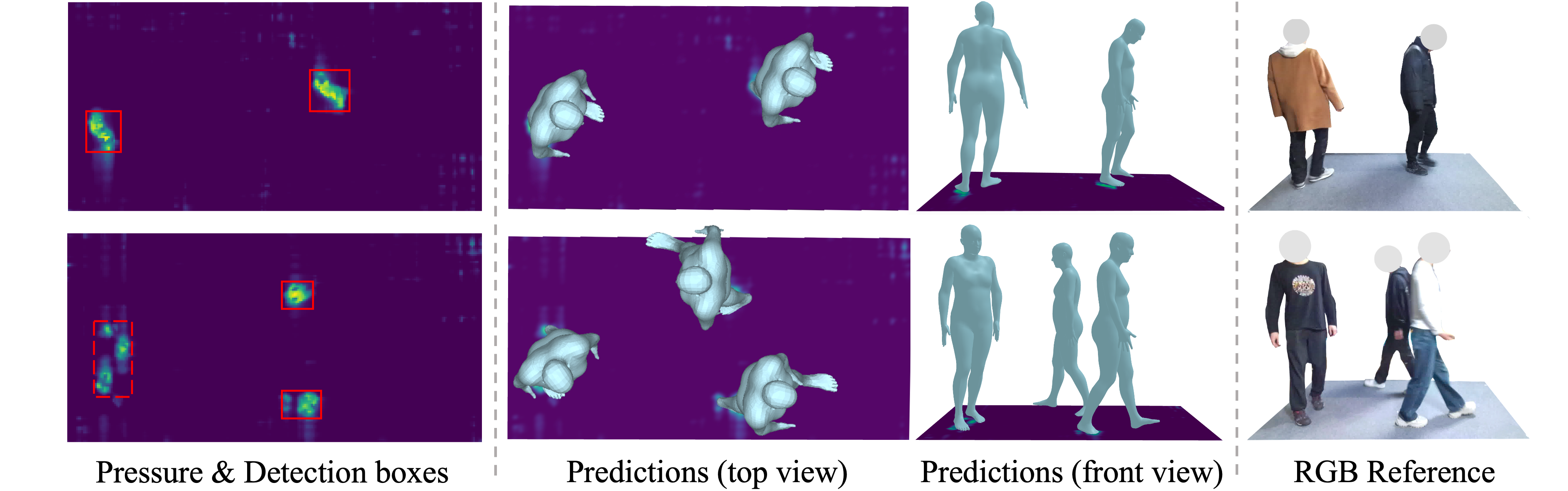} 
\caption{Multi-person human mesh recovery via tactile mat.}
\label{results}
\end{figure*}

\begin{table*}[ht]
    \centering
    \begin{tabular}{l|ccccccccc}
        \hline
        Methods & MOTA $\uparrow$ & MOTP $\uparrow$ & FN $\downarrow$ & FP $\downarrow$ & Frag $\downarrow$ & IDF1 $\uparrow$ & ID sw $\downarrow$ \\
        \hline
        ByteTrack & 66.1\% & 79.1\% & 35207 & 22769 & 17616 & 4.8\% & 14453 \\
        BoT-SORT & 82.5\% & 87.7\% & 17119 & 9020 & 12038 & 6.2\% & 10494 \\
        \textbf{Ours} & \textbf{93.6}\% & \textbf{94.8}\% & \textbf{7317} & \textbf{6764} & \textbf{5270} & \textbf{63.1}\% & \textbf{437} \\
        \hline
    \end{tabular}
    \caption{Footprint tracking performance on MIP.}
    \label{track}
\end{table*}

\section{Evaluation}
To comprehensively evaluate our proposed pipeline, we conduct experiments focusing on both the performance of pressure footprint tracking and the integrated end-to-end human mesh recovery effect.

\textbf{Metrics}
To evaluate different aspects of the tracking performance, we use the CLEARMOT metrics~\cite{CLEARMetrics}, including MOTA (Multi-Object Tracking Accuracy), MOTP (Multi-Object Tracking Precision), FN (number of missed detections), FP (number of false detections), Frag (number of fragmentations where a track is interrupted by miss detection), IDF1 (the harmonic mean of ID precision) and ID sw (number of times an ID switches to a different previously tracked object). 
To evaluate the accuracy of human mesh recovery, we use the following metrics: MPJPE (Mean Per Joint Position Error after pelvis alignment), PA-MPJPE (Procrustes-aligned Mean Per Joint Position Error), and PVE (Per Vertex Error). We compute Accel (Acceleration in~$m/s^2$) to measure the inter-frame smoothness of the reconstructed motion.
Furthermore, we assess motion reconstruction and trajectory estimation accuracy in the world-frame. Following previous work~\cite{shin2024wham}, we split sequences into segments of 100 frames and align each segment with the ground truth using the first two frames (W-MPJPE$_{100}$) or the entire segment (WA-MPJPE$_{100}$). we utilize the GMPJPE (Global Mean Per Joint Position Error), RTE (Root Translation Error in~$\%$) over the entire trajectory, and Jitter (jitter of the motion in the world coordinate system in ~$10m/s^3$). All other metrics are in~$mm$.

\subsection{Footprint Tracking}
Effective tracking is crucial for subsequent human mesh recovery and trajectory estimation, as it directly impacts the identity consistency of the input data. 
Therefore, we quantify the algorithm's ability to accurately differentiate and consistently track the footprints of various individuals in complex multi-person interaction scenarios.
We evaluate our UoE-based pressure footprint tracking method, PressTrack, against ByteTrack~\cite{ByteTrack} and BoT-SORT~\cite{aharon2022bot} on MIP, as these two are among the most influential and widely referenced benchmarks in visual tracking. All trackers employ the high-performance detector YOLOv11, pre-trained with 92.2\% recall and 93.6\% precision. 

As shown in Table~\ref{track}, our method achieves a remarkable MOTA of 93.6\% and a MOTP of 94.8\%, significantly surpassing ByteTrack and SORT in tracking accuracy and localization precision. 
It also excels in maintaining identity consistency and trajectory continuity, with a marked reduction in ID switches.
To contextualize this, our dataset comprises 86 trajectories, averaging 2660 frames each, with 437 total ID switches, implying about one ID switch per 523 frames.
This error is within the acceptable range for our human mesh recovery task (quantitative results are presented in the next subsection), which employs 16-frame continuous sequences as a sliding window.
The overall performance indicates that our method effectively leverages the unique dynamics of pressure footprints to achieve robust and consistent tracking performance.

\begin{figure*}[t]
\centering
\includegraphics[width=0.95\textwidth]{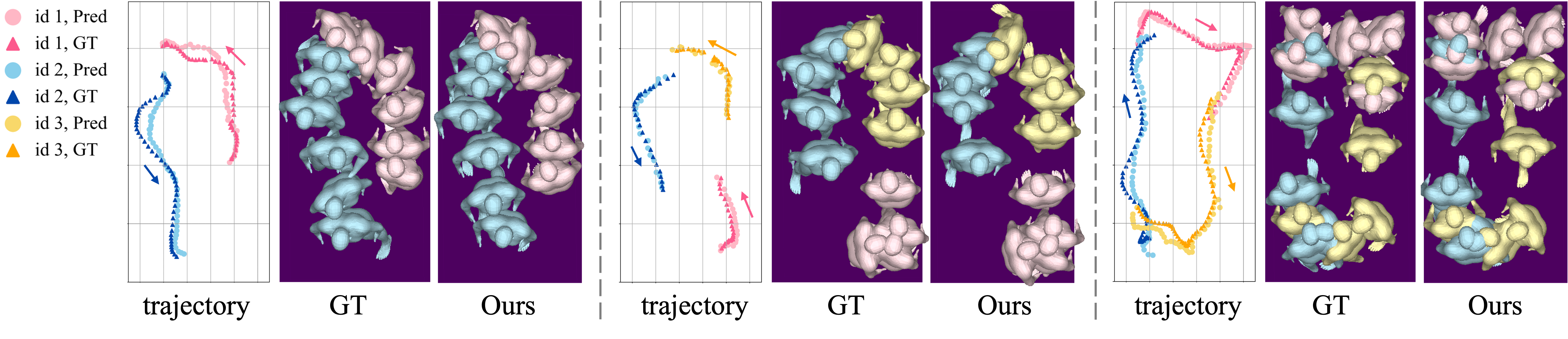} 
\caption{Qualitative results for global trajectory estimation using tracked detections. Our pipeline achieves excellent end-to-end human pose and trajectory estimation.}
\label{trajectory}
\end{figure*}

\begin{table*}[ht]
    \centering
    \begin{tabular}{c|c|cccccccccc}
        \hline
        Methods & Dataset Mode & M. $\downarrow$ & PA-M. $\downarrow$ & PVE $\downarrow$ & Acc. $\downarrow$ & W-M. $\downarrow$ & WA-M. $\downarrow$ & GM. $\downarrow$ & RTE $\downarrow$ & Jit. $\downarrow$\\
        \hline
        \multirow{2}{*}{GT Dets.} 
        & Unseen Sequence & \textbf{81.8} & \textbf{46.1} & \textbf{115.7} & \textbf{4.9} & \textbf{324.3} & \textbf{90.9} & \textbf{99.4} & \textbf{0.62} & \textbf{12.3}\\ 
        & Unseen Subject & 92.2 & 43.1 & 132.9 & 5.2 & 405.8 & 100.8 & 112.8 & 0.13 & 12.8\\
        \hline 
        \multirow{2}{*}{Tracked Dets.}
        & Unseen Sequence & \textbf{89.2} & \textbf{48.8} & \textbf{134.4} & \textbf{8.9} & \textbf{364.0} & \textbf{112.6} & \textbf{118.3} & \textbf{0.96} & \textbf{23.2}\\
        & Unseen Subject & 96.8 & 44.3 & 145.3 & 8.9 & 426.1 & 115.0 & 125.0 & 0.18 & 23.1\\
        \hline
    \end{tabular}
    \caption{Overall Performance of PressTrack-HMR. GT Dets.: Ground-truth Detections, Tracked Dets.: Tracked Detections, M.: MPJPE, PA-M.: PA-MPJPE, Acc.: Accel, W-M.: W-MPJPE$_{100}$, WA-M.: WA-MPJPE$_{100}$, GM.: GMPJPE., Jit.: Jitter.} 
    \label{hmr_evaluation}
\end{table*}

\begin{table}[ht]
    \centering
    \begin{tabular}{l|c|c|c|c|c|c}
        \hline
        Seq. length& 1 & 4 & 8 & 12 & 16 & 32\\
        \hline
        GT Dets.& 94.2 & 88.7 & 86.6 & 83.8 & \textbf{81.8} & 82.1\\
        Tracked Dets.& 96.2 & 92.3 & 90.7 & 89.9 & \textbf{89.2} & 89.7\\
        \hline
    \end{tabular}
    \caption{MPJPE across different sequence length.}
    \label{sequence-lenth}
\end{table}

\subsection{Human Mesh Recovery}
To evaluate the performance of our Human Mesh Recovery module, we employ two distinct dataset splitting strategies: Unseen Sequence Split and Unseen Subject Split. The Unseen Sequence Split allocates 80\% of each recording group's data for training and 20\% for validation and testing, aiming to assess the model's ability to generalize to unseen temporal sequences from known individuals. The Unseen Subject Split uses data from two randomly selected volunteers exclusively for testing, with all other groups for training and validation, thereby evaluating the pipeline's generalization to entirely novel individuals. For all training phases, ground-truth detection boxes serve as input.

Table~\ref{hmr_evaluation} presents the quantitative results of our pipeline. This shows a clear distinction between the performance of the isolated HMR module and the combined end-to-end effect of the entire pipeline, by utilizing either ground-truth detection boxes or tracked detection boxes as input during testing, respectively.
In the unseen-sequence setting, evaluating the HMR module with ground-truth detections yields an MPJPE of 81.8~$mm$ and a WA-MPJPE$_{100}$ of 90.9~$mm$. These results highlight the excellent regression capabilities of our HMR module for both human pose and shape, as well as global trajectory. For the end-to-end evaluation, which leverages tracked detections as input, the MPJPE is 89.2~$mm$ and WA-MPJPE$_{100}$ is 112.6~$mm$, which still represents a favorable outcome. The observed increase in error compared to the isolated HMR module is attributed to the cascading errors originating from the PressTrack module, as the HMR module's input within the complete pipeline directly stems from its tracking outputs, thereby incorporating the errors from the tracking stage.
In the unseen-subject setting, the end-to-end results show only a 7.6~$mm$ MPJPE increase and a 2.4~$mm$ WA-MPJPE$_{100}$ increase compared to the unseen-sequence setting. This indicates that the model still achieves exceptional results for users who have not been previously encountered, highlighting its robust generalization capability to novel individuals.

\textbf{Ablation Study} An ablation study on the choice of sequence length further validates the necessity and benefits of incorporating temporal information through the PressTrack module.
As shown in Table~\ref{sequence-lenth}, for the isolated HMR module (using ground-truth detections), the MPJPE decreases from sequence length 1 to its minimum at sequence length 16 with a 12.37~$mm$ reduction. This indicates that longer input sequences enhance temporal perception, leading to more accurate and smoother reconstructions. Furthermore, this advantage diminishes and errors may increase when sequences become excessively long due to limited temporal correlation, such as at a sequence length of 32. For the end-to-end pipeline (using tracked detections), where accumulated tracking errors are a factor, the MPJPE also minimizes at sequence length 16. However, its 6.99~$mm$ reduction from sequence length 1 is less pronounced than that observed with ground-truth detections, as tracking errors tend to accumulate with increasing sequence duration.

\section{Conclusion}
In this paper, we introduce PressTrack-HMR, a top-down pipeline that recovers multi-person global human meshes solely from pressure signals. We also build the MIP dataset to facilitate the exploration of pressure signals in multi-person scenarios. Extensive experiments demonstrate our approach's capability to generate plausible multi-person human meshes and global trajectories, offering a flexible and scalable solution for RGB-constrained multi-person motion capture.

We have observed that the performance gap between using ground-truth detections and tracked detections is notable. For future work, we will analyze the impact of different types of tracking errors (e.g., ID switches vs. localization jitter) on HMR performance, aiming to reduce the cascaded errors of the tracking module.

\bibliography{My_reference}

\begin{thebibliography}{35}
\providecommand{\natexlab}[1]{#1}

\bibitem[{Aharon, Orfaig, and Bobrovsky(2022)}]{aharon2022bot}
Aharon, N.; Orfaig, R.; and Bobrovsky, B.-Z. 2022.
\newblock BoT-SORT: Robust associations multi-pedestrian tracking.
\newblock \emph{arXiv preprint arXiv:2206.14651}.

\bibitem[{Arnab, Doersch, and Zisserman(2019)}]{exploiting}
Arnab, A.; Doersch, C.; and Zisserman, A. 2019.
\newblock Exploiting temporal context for 3D human pose estimation in the wild.
\newblock In \emph{Proceedings of the IEEE/CVF Conference on Computer Vision and Pattern Recognition}, 3395--3404.

\bibitem[{Bernardin and Stiefelhagen(2008)}]{CLEARMetrics}
Bernardin, K.; and Stiefelhagen, R. 2008.
\newblock Evaluating multiple object tracking performance: the clear mot metrics.
\newblock \emph{EURASIP Journal on Image and Video Processing}, 2008(1): 246309.

\bibitem[{Bewley et~al.(2016)Bewley, Ge, Ott, Ramos, and Upcroft}]{SORT}
Bewley, A.; Ge, Z.; Ott, L.; Ramos, F.; and Upcroft, B. 2016.
\newblock Simple online and realtime tracking.
\newblock In \emph{2016 IEEE International Conference on Image Processing (ICIP)}, 3464--3468. IEEE.

\bibitem[{Bogo et~al.(2016)Bogo, Kanazawa, Lassner, Gehler, Romero, and Black}]{smplify}
Bogo, F.; Kanazawa, A.; Lassner, C.; Gehler, P.; Romero, J.; and Black, M.~J. 2016.
\newblock Keep it {SMPL}: Automatic estimation of 3D human pose and shape from a single image.
\newblock In \emph{Computer Vision--ECCV 2016: 14th European Conference, Amsterdam, The Netherlands, October 11-14, 2016, Proceedings, Part V 14}, 561--578. Springer.

\bibitem[{Chen et~al.(2024)Chen, Hu, Song, Liu, Torralba, and Matusik}]{cavatar}
Chen, W.; Hu, Y.; Song, W.; Liu, Y.; Torralba, A.; and Matusik, W. 2024.
\newblock CAvatar: Real-time Human Activity Mesh Reconstruction via Tactile Carpets.
\newblock \emph{Proc. ACM Interact. Mob. Wearable Ubiquitous Technol.}, 7(4).

\bibitem[{Cheng et~al.(2024)Cheng, Song, Ge, Liu, Wang, and Shan}]{Cheng2024YOLOWorld}
Cheng, T.; Song, L.; Ge, Y.; Liu, W.; Wang, X.; and Shan, Y. 2024.
\newblock YOLO-World: Real-Time Open-Vocabulary Object Detection.
\newblock In \emph{Proc. IEEE Conf. Computer Vision and Pattern Recognition (CVPR)}.

\bibitem[{Chu et~al.(2023)Chu, Wang, You, Ling, and Liu}]{transmot}
Chu, P.; Wang, J.; You, Q.; Ling, H.; and Liu, Z. 2023.
\newblock TransMOT: Spatial-Temporal Graph Transformer for Multiple Object Tracking.
\newblock In \emph{2023 IEEE/CVF Winter Conference on Applications of Computer Vision (WACV)}, 4859--4869.

\bibitem[{Clever et~al.(2020)Clever, Erickson, Kapusta, Turk, Liu, and Kemp}]{Clever2020Bodies}
Clever, H.~M.; Erickson, Z.; Kapusta, A.; Turk, G.; Liu, K.; and Kemp, C.~C. 2020.
\newblock Bodies at rest: 3d human pose and shape estimation from a pressure image using synthetic data.
\newblock In \emph{Proceedings of the IEEE/CVF Conference on Computer Vision and Pattern Recognition ({CVPR})}, 6215--6224.

\bibitem[{He et~al.(2016)He, Zhang, Ren, and Sun}]{resnet}
He, K.; Zhang, X.; Ren, S.; and Sun, J. 2016.
\newblock Deep Residual Learning for Image Recognition.
\newblock In \emph{Proceedings of the IEEE Conference on Computer Vision and Pattern Recognition (CVPR)}.

\bibitem[{Kalman(1960)}]{Kalman}
Kalman, R.~E. 1960.
\newblock A New Approach to Linear Filtering and Prediction Problems.
\newblock \emph{Journal of Basic Engineering}, 82(1): 35--45.

\bibitem[{Kanazawa et~al.(2018)Kanazawa, Black, Jacobs, and Malik}]{hmr}
Kanazawa, A.; Black, M.~J.; Jacobs, D.~W.; and Malik, J. 2018.
\newblock End-to-end recovery of human shape and pose.
\newblock In \emph{Proceedings of the IEEE Conference on Computer Vision and Pattern Recognition}, 7122--7131.

\bibitem[{Khanam and Hussain(2024)}]{2024yolov11}
Khanam, R.; and Hussain, M. 2024.
\newblock YOLOv11: An Overview of the Key Architectural Enhancements.
\newblock arXiv:2410.17725.

\bibitem[{Kuhn(1955)}]{Hungarian}
Kuhn, H.~W. 1955.
\newblock The Hungarian method for the assignment problem.
\newblock \emph{Naval Research Logistics Quarterly}, 2(1-2): 83--97.

\bibitem[{Loper et~al.(2015)Loper, Mahmood, Romero, Pons-Moll, and Black}]{smpl}
Loper, M.; Mahmood, N.; Romero, J.; Pons-Moll, G.; and Black, M.~J. 2015.
\newblock SMPL: A Skinned Multi-Person Linear Model.
\newblock \emph{ACM Transactions on Graphics}, 34(6).

\bibitem[{Luo et~al.(2021)Luo, Li, Foshey, Shou, Sharma, Palacios, Torralba, and Matusik}]{Luo_IntelligentCarpet}
Luo, Y.; Li, Y.; Foshey, M.; Shou, W.; Sharma, P.; Palacios, T.; Torralba, A.; and Matusik, W. 2021.
\newblock Intelligent Carpet: Inferring 3D Human Pose From Tactile Signals.
\newblock In \emph{Proceedings of the IEEE/CVF Conference on Computer Vision and Pattern Recognition (CVPR)}, 11255--11265.

\bibitem[{Pavlakos et~al.(2019)Pavlakos, Choutas, Ghorbani, Bolkart, Osman, Tzionas, and Black}]{pavlakos2019expressive}
Pavlakos, G.; Choutas, V.; Ghorbani, N.; Bolkart, T.; Osman, A.~A.; Tzionas, D.; and Black, M.~J. 2019.
\newblock Expressive body capture: 3d hands, face, and body from a single image.
\newblock In \emph{Proceedings of the IEEE/CVF Conference on Computer Vision and Pattern Recognition}, 10975--10985.

\bibitem[{Redmon and Farhadi(2018)}]{yolov3}
Redmon, J.; and Farhadi, A. 2018.
\newblock YOLOv3: An Incremental Improvement.
\newblock arXiv:1804.02767.

\bibitem[{Ren et~al.(2025)Ren, Lu, Huang, Zhao, Zhang, Yu, Shen, and Cao}]{Ren_MotionPRO}
Ren, S.; Lu, Y.; Huang, J.; Zhao, J.; Zhang, H.; Yu, T.; Shen, Q.; and Cao, X. 2025.
\newblock MotionPRO: Exploring the Role of Pressure in Human MoCap and Beyond.
\newblock In \emph{Proceedings of the Computer Vision and Pattern Recognition Conference (CVPR)}, 27760--27770.

\bibitem[{Shin et~al.(2024)Shin, Kim, Halilaj, and Black}]{shin2024wham}
Shin, S.; Kim, J.; Halilaj, E.; and Black, M.~J. 2024.
\newblock {WHAM}: Reconstructing world-grounded humans with accurate 3d motion.
\newblock In \emph{Proceedings of the IEEE/CVF Conference on Computer Vision and Pattern Recognition ({CVPR})}, 2070--2080.

\bibitem[{Shuai et~al.(2022)Shuai, Geng, Fang, Peng, Shen, Zhou, and Bao}]{easymocap}
Shuai, Q.; Geng, C.; Fang, Q.; Peng, S.; Shen, W.; Zhou, X.; and Bao, H. 2022.
\newblock Novel View Synthesis of Human Interactions from Sparse Multi-view Videos.
\newblock In \emph{ACM SIGGRAPH 2022 Conference Proceedings}, SIGGRAPH '22. New York, NY, USA: Association for Computing Machinery.

\bibitem[{Tripathi et~al.(2023)Tripathi, M{\"u}ller, Huang, Taheri, Black, and Tzionas}]{moyo}
Tripathi, S.; M{\"u}ller, L.; Huang, C.-H.~P.; Taheri, O.; Black, M.~J.; and Tzionas, D. 2023.
\newblock 3D human pose estimation via intuitive physics.
\newblock In \emph{Proceedings of the IEEE/CVF Conference on Computer Vision and Pattern Recognition}, 4713--4725.

\bibitem[{Van~Wouwe et~al.(2024)Van~Wouwe, Lee, Falisse, Delp, and Liu}]{VanWouwe2024Diffusionposer}
Van~Wouwe, T.; Lee, S.; Falisse, A.; Delp, S.; and Liu, C.~K. 2024.
\newblock Diffusionposer: Real-time human motion reconstruction from arbitrary sparse sensors using autoregressive diffusion.
\newblock In \emph{Proceedings of the IEEE/CVF Conference on Computer Vision and Pattern Recognition ({CVPR})}, 2513--2523.

\bibitem[{Vaswani et~al.(2017)Vaswani, Shazeer, Parmar, Uszkoreit, Jones, Gomez, Kaiser, and Polosukhin}]{vaswani2017attention}
Vaswani, A.; Shazeer, N.; Parmar, N.; Uszkoreit, J.; Jones, L.; Gomez, A.; Kaiser, L.; and Polosukhin, I. 2017.
\newblock Attention Is All You Need.
\newblock In \emph{Advances in Neural Information Processing Systems}, volume~30.

\bibitem[{Wan et~al.(2025)Wan, Wu, Xie, Niu, and Cai}]{wan2025bed}
Wan, Q.; Wu, Z.; Xie, F.; Niu, M.; and Cai, X. 2025.
\newblock In-bed Pressure Image-supported Diffusion for 3D Human Mesh Recovery.
\newblock In \emph{2025 IEEE International Conference on Pervasive Computing and Communications (PerCom)}, 89--98. IEEE.

\bibitem[{Wang, Bochkovskiy, and Liao(2022)}]{yolov7}
Wang, C.-Y.; Bochkovskiy, A.; and Liao, H.-Y.~M. 2022.
\newblock YOLOv7: Trainable bag-of-freebies sets new state-of-the-art for real-time object detectors.
\newblock arXiv:2207.02696.

\bibitem[{Wang et~al.(2025)Wang, Sun, Patel, Daniilidis, Black, and Kocabas}]{PromptHMR}
Wang, Y.; Sun, Y.; Patel, P.; Daniilidis, K.; Black, M.~J.; and Kocabas, M. 2025.
\newblock PromptHMR: Promptable Human Mesh Recovery.
\newblock arXiv:2504.06397.

\bibitem[{Wojke, Bewley, and Paulus(2017)}]{DeepSORT}
Wojke, N.; Bewley, A.; and Paulus, D. 2017.
\newblock Simple online and realtime tracking with a deep association metric.
\newblock In \emph{2017 IEEE International Conference on Image Processing (ICIP)}, 3645--3649.

\bibitem[{Wu et~al.(2024)Wu, Xie, Fang, Liang, Wan, Xiong, and Cai}]{pimesh}
Wu, Z.; Xie, F.; Fang, Y.; Liang, Z.; Wan, Q.; Xiong, Y.; and Cai, X. 2024.
\newblock Seeing through the Tactile: 3D Human Shape Estimation from Temporal In-Bed Pressure Images.
\newblock \emph{Proc. ACM Interact. Mob. Wearable Ubiquitous Technol.}, 8(2).

\bibitem[{Wu et~al.(2025)Wu, Xiong, Niu, Xie, Wan, Ying, Liu, and Cai}]{wu2025pihmr}
Wu, Z.; Xiong, Y.; Niu, M.; Xie, F.; Wan, Q.; Ying, Q.; Liu, B.; and Cai, X. 2025.
\newblock PI-HMR: Towards Robust In-bed Temporal Human Shape Reconstruction with Contact Pressure Sensing.
\newblock arXiv:2503.00068.

\bibitem[{Yuan et~al.(2022)Yuan, Iqbal, Molchanov, Kitani, and Kautz}]{yuan2022glamr}
Yuan, Y.; Iqbal, U.; Molchanov, P.; Kitani, K.; and Kautz, J. 2022.
\newblock {GLAMR}: Global occlusion-aware human mesh recovery with dynamic cameras.
\newblock In \emph{Proceedings of the IEEE/CVF Conference on Computer Vision and Pattern Recognition ({CVPR})}, 11038--11049.

\bibitem[{Zhang et~al.(2024{\natexlab{a}})Zhang, Liang, Wu, Xie, Xu, Wu, and Cai}]{Zhang2024Garment}
Zhang, D.; Liang, Z.; Wu, Y.; Xie, F.; Xu, G.; Wu, Z.; and Cai, X. 2024{\natexlab{a}}.
\newblock Learn to infer human poses using a full-body pressure sensing garment.
\newblock \emph{IEEE Sensors Journal}.

\bibitem[{Zhang et~al.(2024{\natexlab{b}})Zhang, Ren, Yuan, Zhao, Li, Sun, Liang, Yu, Shen, and Cao}]{Zhang2024MMVP}
Zhang, H.; Ren, S.; Yuan, H.; Zhao, J.; Li, F.; Sun, S.; Liang, Z.; Yu, T.; Shen, Q.; and Cao, X. 2024{\natexlab{b}}.
\newblock {MMVP}: A multimodal mocap dataset with vision and pressure sensors.
\newblock In \emph{Proceedings of the IEEE/CVF Conference on Computer Vision and Pattern Recognition ({CVPR})}, 21842--21852.

\bibitem[{Zhang et~al.(2022)Zhang, Sun, Jiang, Yu, Weng, Yuan, Luo, Liu, and Wang}]{ByteTrack}
Zhang, Y.; Sun, P.; Jiang, Y.; Yu, D.; Weng, F.; Yuan, Z.; Luo, P.; Liu, W.; and Wang, X. 2022.
\newblock {ByteTrack}: {Multi}-object {Tracking} by {Associating} {Every} {Detection} {Box}.
\newblock In \emph{Computer {Vision} – {ECCV} 2022}, volume 13682, 1--21. Cham: Springer Nature Switzerland.

\bibitem[{Zhou et~al.(2023)Zhou, Geissler, Faulhaber, Gleiss, Zahn, Ray, Gamarra, Rey, Suh, Bian et~al.}]{Zhou2023Mocapose}
Zhou, B.; Geissler, D.; Faulhaber, M.; Gleiss, C.~E.; Zahn, E.~F.; Ray, L. S.~S.; Gamarra, D.; Rey, V.~F.; Suh, S.; Bian, S.; et~al. 2023.
\newblock {Mocapose}: Motion capturing with textile-integrated capacitive sensors in loose-fitting smart garments.
\newblock \emph{Proceedings of the ACM on Interactive, Mobile, Wearable and Ubiquitous Technologies}, 7(1): 1--40.

\end{thebibliography}
\end{document}